\documentclass[runningheads]{llncs}
\usepackage{graphicx}
\usepackage{epsfig, subfigure}
\usepackage{amsmath,amssymb}
\usepackage{color}
\usepackage{amsfonts}
\newcommand{\questionnet}{\mbox{\sc Dqa-Net}}  
\newcommand{\lstmnetwork}{\mbox{\sc Dsdp-Net}}  
\newcommand{\datasetname}{\mbox{\sc AI2D}}  

\usepackage[width=122mm,left=12mm,paperwidth=146mm,height=193mm,top=12mm,paperheight=217mm]{geometry}
\begin{document}
\pagestyle{headings}
\mainmatter

\title{A Diagram Is Worth A Dozen Images}

\titlerunning{A Diagram Is Worth A Dozen Images}
\authorrunning{Kembhavi et al.}

\author{Aniruddha Kembhavi$^\dag$, Mike Salvato$^{\dag}$\thanks{These authors contributed equally to this work.}, Eric Kolve$^{\dag}$$^\star$, Minjoon Seo$^\S$,\\Hannaneh Hajishirzi$^\S$, Ali Farhadi$^{\dag \S}$}
\institute{$^\dag$Allen Institute for Artificial Intelligence, $^\S$University of Washington}

\maketitle
\begin{abstract}
Diagrams are common tools for representing complex concepts, relationships and events, often when it would be difficult to portray the same information with natural images. Understanding natural images has been extensively studied in computer vision, while diagram understanding has received little attention. In this paper, we study the problem of diagram interpretation and reasoning, the challenging task of identifying the structure of a diagram and the semantics of its constituents and their relationships. We introduce Diagram Parse Graphs (DPG) as our representation to model the structure of diagrams. We define syntactic parsing of diagrams as learning to infer DPGs for diagrams and study semantic interpretation and reasoning of diagrams in the context of diagram question answering. We devise an LSTM-based method  for syntactic parsing of diagrams and  introduce a DPG-based attention model for diagram question answering. We compile a new dataset of diagrams with exhaustive annotations of constituents and relationships for over 5,000 diagrams and 15,000 questions and answers. Our results show the significance of our models for syntactic parsing and question answering in diagrams using DPGs. 
\end{abstract}

\section{Introduction}
\begin{figure}[t]
\centering
\includegraphics[width=27pc]{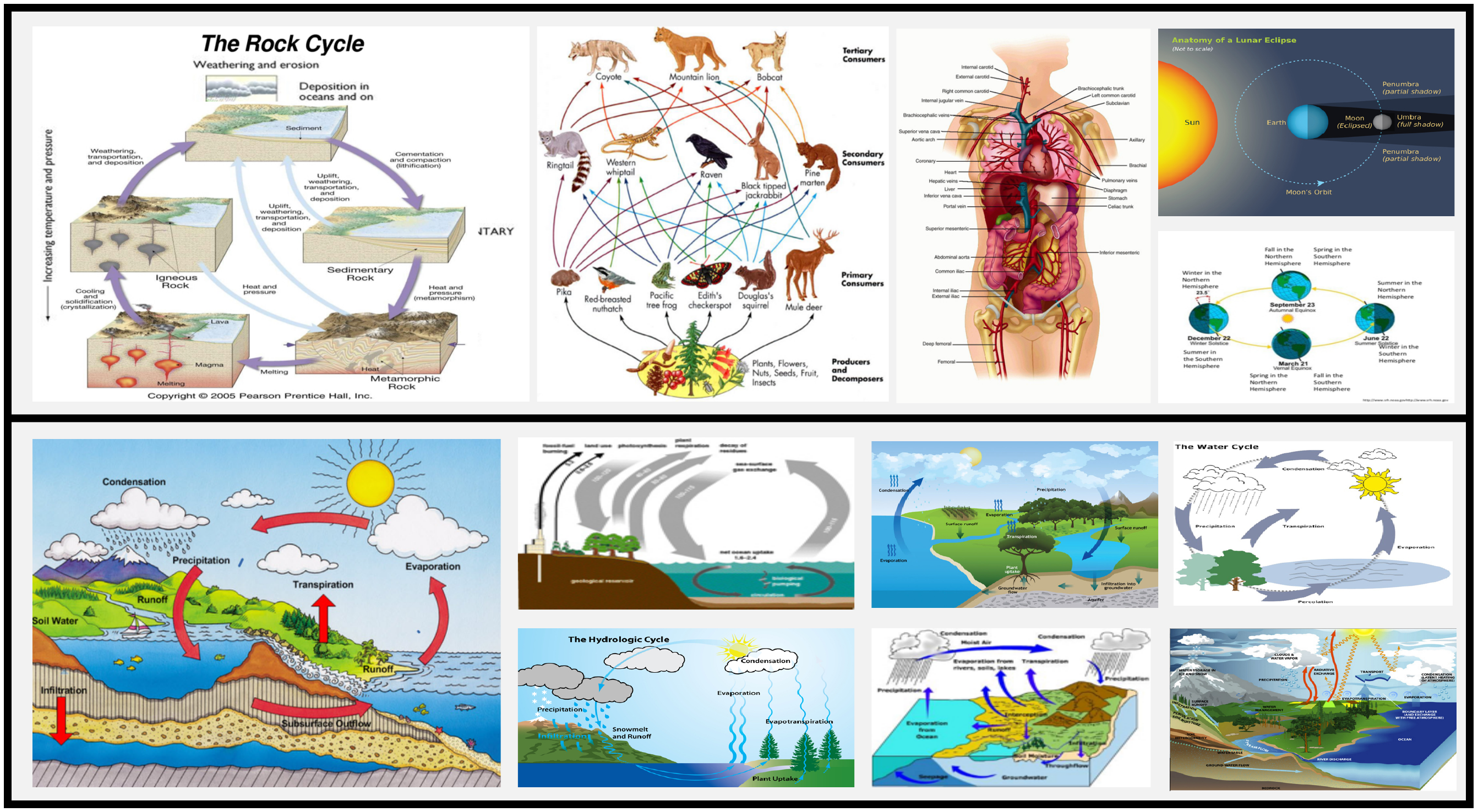}\vspace{-.2cm}
\caption{\small The space of visual illustrations is very rich and diverse. The top palette shows the inter class variability for diagrams in our new diagram  dataset, \datasetname. The bottom palette shows the intra-class variation for the Water Cycles category.}\vspace{-.2cm}
\label{fig:diversity}
\end{figure}

For thousands of years visual illustrations have been used to depict the lives of people, animals,  their environment, and major events. Archaeological discoveries have unearthed cave paintings showing lucid representations of hunting, religious rites, communal dancing, burial, etc. 
From ancient rock carvings and maps, to modern info-graphics and 3-D visualizations, to diagrams in science textbooks, the set of visual illustrations  is very large, diverse and ever growing, constituting a considerable portion of visual data. These  illustrations  often represent complex concepts, such as events or systems, that are otherwise difficult to portray in a few sentences of text or a natural image (Figure~\ref{fig:diversity}). 

While understanding natural images has been a major area of research in computer vision, understanding rich visual illustrations has received scant attention.   From a computer vision perspective, these illustrations are inherently different from natural images and offer a unique and interesting set of problems. Since they are purposefully designed to express information, they typically suppress irrelevant signals such as background clutter, intricate textures and shading nuances. This often makes the detection and recognition of individual elements inherently different than their counterparts, objects, in natural images. On the other hand, visual illustrations may depict complex phenomena and higher-order relations between objects (such as temporal transitions, phase transformations and inter object dependencies) that go well beyond what a single natural image can convey. For instance, one might struggle to find natural images that compactly represent the phenomena seen in some grade school science diagrams, as shown in Figure~\ref{fig:diversity}. In this paper, we define the problem of understanding visual illustrations by identifying visual entities and their relations as well as establishing semantic correspondences to real-world concepts.

The characteristics of visual illustrations also afford opportunities for deeper reasoning than provided by natural images. Consider the food web in Figure~\ref{fig:diversity}, which represents several relations such as foxes eating rabbits and rabbits eating plants. One can further reason about higher order relations between entities such as the effect on the population of foxes caused by a reduction in the population of plants. Similarly, consider the myriad of phenomena displayed in a single water cycle diagram in Figure~\ref{fig:diversity}. Some of these phenomena are shown to occur on the surface of the earth while others occur either above or below the surface. The main components of the cycle (e.g., evaporation and condensation) are labeled and the flow of water is displayed using arrows. Reasoning about these objects and their interactions in such rich scenes provides many exciting research challenges.

In this paper, we address the problem of {\it diagram} interpretation and reasoning in the context of science diagrams, defined as the two tasks of \textit{Syntactic parsing} and \textit{Semantic interpretation}. \textit{Syntactic parsing} involves detecting and recognizing constituents and their syntactic relationships in a diagram. This is most analogous to the problem of scene parsing in natural images. The wide variety of diagrams as well as large intra-class variation (Figure~\ref{fig:diversity} shows several varied images depicting a water cycle) make this step very challenging.  \textit{Semantic interpretation} is the task of  mapping constituents and their relationships  to semantic entities and events (real-world concepts). This is a challenging task given the inherent ambiguities in the mapping functions. For example, an arrow in a food chain diagram typically corresponds to the concept of \textit{consumption}, arrows in water cycles typically refer to {\it phase changes}, and arrows in a planetary diagram often refers to {\it rotatory motion}.

We introduce a  representation to encode diagram constituents and their relationships in a graph, called diagram parse graphs (DPG) (example DPGs are shown in Figure~\ref{fig:dpgs}).
The problem of syntactic parsing of diagrams is formulated as the task of learning to infer the DPG that best explains a diagram.  We introduce a Deep Sequential Diagram Parser Network (\lstmnetwork) that learns to sequentially add relationships and their constituents to form DPGs, using Long Short Term Memory (LSTM) networks. The problem of semantically interpreting a diagram and reasoning about the constituents and their relationships is studied in the context of diagram question answering.  We present a neural network architecture (called \questionnet) that learns to attend to useful relations in a DPG given a question about the diagram.  

We compile a dataset named AI2 Diagrams (\datasetname)  of over 5000 grade school science diagrams with over 150000 rich annotations, their ground truth syntactic parses, and more than 15000 corresponding multiple choice questions. Our experimental results show that the proposed \lstmnetwork\ for syntactic parsing outperforms several baseline methods. Moreover, we show that the proposed approach of incorporating diagram relations into question answering outperforms standard visual question answering methods. 

Our contributions include: (a) We present two new tasks of diagram interpretation and reasoning, (b) we introduce the DPG representation to encode diagram parses and introduce a model that learns to map diagrams into DPGs, (c) we introduce a model for diagram question answering that learns the attention of questions into DPGs and (d) we present a new dataset to evaluate the above models with baselines.

\section{Background}

\noindent{\bf Understanding diagrams.}
The problem of understanding diagrams received a fair amount of interest ~\cite{Srihari1994ComputationalMF,Ferguson98tellingjuxtapositions,Watanabe1998DiagramUU,ogorman1997dia} in the 80's and 90's. However, many of these techniques either used hand written rules, assumed that the visual primitives were manually identified or worked on a specific set of diagrams. More recently, Futrelle et al.\cite{Futrelle2003ExtractionLA} proposed methods to analyze graphs and finite automata sketches but only worked with vector representations of these diagrams. Recently, Seo et al. \cite{Seo2014DiagramUI} proposed a method for understanding diagrams in geometry questions that identifies visual elements in a diagram while maximizing agreement between textual and visual data. In contrast to these past approaches, we propose a unified approach to diagram understanding that builds upon the representational language of graphic representations proposed by Engelhardt \cite{von2002language} and works on a diverse set of diagrams. 

The domain of abstract images has also received a considerable amount of interest over the past couple of years~\cite{Zitnick2013BringingSI,Zhang2015YinAY,Vedantam2015LearningCS,Antol2014ZeroShotLV}. While abstract images significantly reduce the noise introduced by low level computer vision modules, thus bringing the semantics of the scene into focus, they still depict real world scenes, and hence differ significantly from diagrams which may depict more complex phenomena.

\vspace{.1cm}
\noindent{\bf Parsing natural images.}
Several approaches to building bottom-up and top-down parsers have been proposed to syntactically parse natural images and videos. These include Bayesian approaches~\cite{Tu2003ImagePU}, And-Or graph structures~\cite{Zhu2006ASG}, stochastic context free grammars~\cite{Martinovic2013BayesianGL}, regular grammars~\cite{Pirsiavash2014ParsingVO}, 3D Geometric Phrases~\cite{Choi2013UnderstandingIS} and a max margin structured prediction framework based on recursive neural networks~\cite{Socher2011ParsingNS}. Inspired by these methods, we adopt a graph based representation for diagrams.

\vspace{.1cm}
\noindent{\bf Answering questions.}
One of relevant tasks in NLP is machine comprehension, which is to answer questions about a reading passage~\cite{WestonBCM15,RichardsonBR13,hermann2015teaching}.
Our QA system is similar to memory networks~\cite{sukhbaatar2015end} in that we use attention mechanism to focus on the best supporting fact among possible candidates.
While these candidates are trivially obtained from the given passage in~\cite{sukhbaatar2015end}, in DQA we obtain them from diagram via its parse graph.  
Recently, answering questions about real images has drawn much attention in both NLP and vision communities~\cite{antol2015vqa,ren2015exploring,ZhuGBF15}.
However, diagram images are vastly different from real images, and so are the corresponding questions. Hence, most QA systems built for real images~\cite{andreas2016learning,noh2015image} cannot be directly used for diagram QA tasks.
\section{The Language of Diagrams}\label{sec:rep}
Much of the existing literature on graphic representations \cite{horn1998visual,card1999readings,twyman1979schema} covers only specific types of graphics or specific aspects of their syntactic structure. More recently, Engelhardt~\cite{von2002language} proposed a coherent framework integrating various structural, semiotic and classification aspects that can be applied to the complete spectrum of graphic representations including diagrams, maps and more complex computer visualizations.
We briefly describe some of his proposed principles below, as they apply to our space of diagrams, but refer the reader to \cite{von2002language} for a more thorough understanding.

A diagram is a composite graphic that consists of a graphic space, a set of constituents,  and a set of relationships involving these constituents. A graphic space may be a metric space, 
distorted metric space (e.g., a solar system diagram) 
 or a non-meaningful space (e.g., a food web). Constituents in a diagram include illustrative elements (e.g., drawings of animals and plants), textual elements, diagrammatic elements (e.g., arrows and grid lines), informative elements (e.g., legends and captions) and decorative elements. Relationships include spatial relations between constituents and their positions in the diagram space, and spatial and attribute-based relations between constituents (e.g., linkage, lineup, variation in color, shape). 
An individual constituent may itself be a composite graphic, rendering this formulation recursive. 

\begin{table}[t]
    \centering
    \begin{small}
    \begin{tabular}{p{12cm}} \hline
    \noindent \textbf{Intra-Object Label ($\mathbb{R}_1$):} A text box naming the entire object.\\
\noindent \textbf{Intra-Object Region Label ($\mathbb{R}_2$):} A text box referring to a region within an object.\\
\noindent \textbf{Intra-Object Linkage ($\mathbb{R}_3$):} A text box referring to a region within an object via an arrow.\\
\noindent \textbf{Inter-Object Linkage ($\mathbb{R}_4$):} Two objects related to one another via an arrow.
\\
\noindent \textbf{Arrow Head Assignment ($\mathbb{R}_5$):} An arrow head associated to an arrow tail.
\\
\noindent \textbf{Arrow Descriptor ($\mathbb{R}_6$):} A text box describing a process that an arrow refers to.
\\
\noindent \textbf{Image Title ($\mathbb{R}_7$):} The title of the entire image.
\\
\noindent \textbf{Image Section Title ($\mathbb{R}_8$):} Text box that serves as a title for a section of the image.
\\
\noindent \textbf{Image Caption ($\mathbb{R}_9$):} A text box that adds information about the entire image, but does not serve as the image title. 
\\
\noindent \textbf{Image Misc ($\mathbb{R}_{10}$):} Decorative elements in the diagram. \\ \hline
    \end{tabular}
    \end{small}
    \caption{\small Different types of relationships in our diagram parse graphs.}
    \label{tab:relations}
\end{table}

\vspace{.1cm}
\noindent {\bf Our Representation: Diagram Parse Graph.} 
We build upon Engelhardt's representation by introducing the concept of \textit{Diagrammatic Objects} in our diagrams, defined as the primary entities being described in the diagram. Examples of objects include animals in the food web, the human body in an anatomy diagram,  and the sun in water cycle (Figure~\ref{fig:diversity}). The relationships within and between objects include intra-object, inter-object, and constituent-space relationships. 
We represent a diagram with a  {\it Diagram Parse Graph} (DPG), in which nodes correspond to {\it constituents} and edges  correspond to {\it relationships} between the constituents. We model four types of constituents:  Blobs (Illustrations), Text Boxes, Arrows, and Arrow Heads.\footnote{Separating arrow heads from arrow tails enables us to represent arrows with a single head, multiple heads or without heads in a uniform way.} We also model ten classes of relationships summarized in Table~\ref{tab:relations}. Figure~\ref{fig:dpgs} shows some examples of DPGs, their different constituents and relationships in our dataset.

\section{ Syntactic Diagram Parsing}\label{sec:syn}
Syntactic diagram parsing is the problem of learning to map diagrams into DPGs. Specifically, the goal is to detect and recognize constituents and their syntactic relationships in a diagram and find the DPG that best explains the diagram.  
In order to form candidate DPGs, we first generate proposals for nodes in the DPG using object detectors built for each constituent category (Section~\ref{sec:proposal1}). We also generate proposals for edges in the DPG by combining proposal nodes using relationship classifiers (Section~\ref{sec:proposal2}). Given sets of noisy node and edge proposals, our method then selects a subset of these to form a DPG by exploiting several local and global cues.

The constituent and relationship proposal generators result in several hundred constituent proposals and several thousand relationship proposals per diagram. These large sets of proposals, the relatively smaller number of true nodes and edges in the truth DPG and the rich nature of the structure of our DPGs, makes the search space for possible parse graphs incredibly large. 
We observe that forming a DPG amounts to choosing a subset of relationships among the proposals. Therefore, we propose a sequential formulation to this task that adds a relationship and its constituents at every step, exploiting local cues  as well as long range global contextual cues using a memory-based model.

\begin{figure}[t]
\centering
\includegraphics[width=27pc]{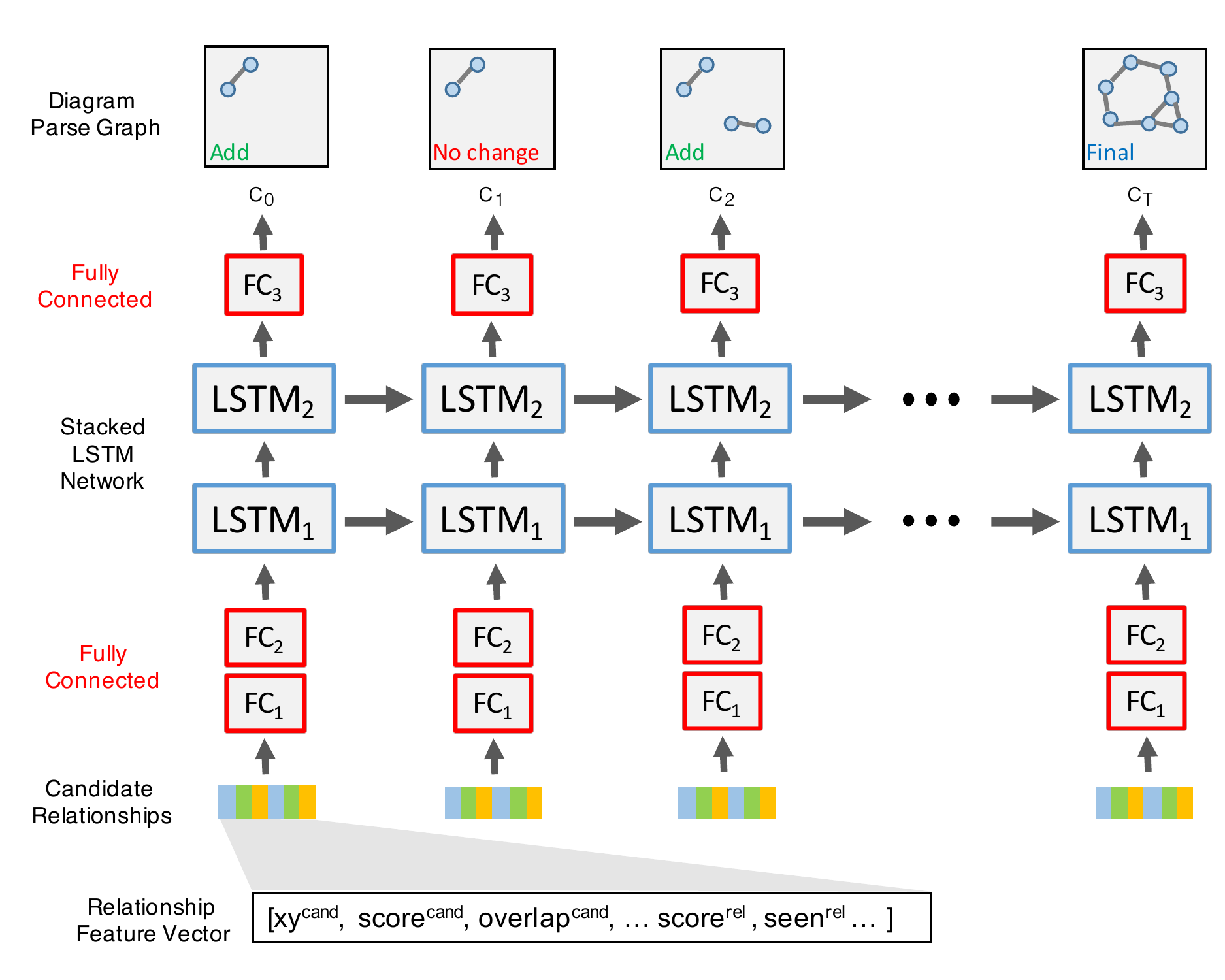}\vspace{-.2cm}
\caption{\small An overview of the \lstmnetwork\ solution to inferring DPGs from diagrams. The LSTM based network exploits global constrains such as overlap, coverage, and layout to select a subset of relations amongst thousands of candidates to construct a DPG.}
\label{fig:lstmNetwork}
\end{figure}

\vspace{.1cm}
 \noindent {\bf Model.} We introduce a Deep Sequential Diagram Parser (\lstmnetwork). Figure~\ref{fig:lstmNetwork} depicts an unrolled illustration of \lstmnetwork. Central to this is a stacked Long Short Term Memory (LSTM) recurrent neural network~\cite{Hochreiter1997LongSM} with fully connected layers used prior to, and after the LSTM. Proposal relationships are then sequentially fed into the network, one at every time step, and the network predicts if this relationship (and its constituents) should be added to the DPG or not. Each relationship in our large candidate set is represented by a feature vector, capturing the spatial layout of its constituents in image space and their detection scores (more details in Section~\ref{sec:lstmexperiment}).  

{\bf Training.} LSTM networks typically require large amounts of training data. We provide training data for the \lstmnetwork\ in the form of sequences of relationships by sampling from training diagrams. For each training diagram, we sample a large number of relationship sequences using sampling without replacement from  thousands of relationship candidates, utilizing the relationship proposal scores as sampling weights. For each sampled sequence, we sequentially label the relationship at every time step by comparing the generated DPG to the groundtruth DPG.\footnote{A relationship labeled with a positive label in one sampled sequence may be labeled with a negative label in another sequence due to the presence of overlapping constituents and relationships in our candidate sets.}

The \lstmnetwork\  is able to model dependencies between nodes and edges selected at different time steps in the sequence. It chooses relationships with large proposal scores but also learns to reject relationships that may lead to a high level of spatial redundancy or an incorrect structure in the layout. It also works well with a variable number of candidate relationships per diagram. Finally, it  learns to stop adding relationships once the entire image space has been covered by the nodes and edges already present in the graph.

{\bf Test.} At test time, relationships in the candidate set are sorted by their proposal scores and presented to the network. Selected relationships are then sequentially added to form the final DPG. 

\section{Semantic Interpretation}\label{sec:semantic}
DPGs represent the syntactic relationships between constituents of a diagram. They, however, do not encode the semantics of constituents and relationships. For example, the corresponding DPG in Figure~\ref{fig:datasetComplexity} indicates that \texttt{tree} and \texttt{mule deer} are related via in Inter-Object Linkage relationship, but it does not represent that the arrow corresponds to {\it consuming}. Constituents and relationships with a similar visual representation may have different semantic meanings in different diagrams. For example, the Inter-Object Linkage relationship can be interpreted as {\it consuming} in food webs and as {\it evaporation} in water cycles. Moreover, diagrams typically depict complex phenomena and reasoning about these phenomena goes beyond the tasks of matching and interpretation. For example, answering the question in Figure~\ref{fig:datasetComplexity} requires parsing the relationship between \texttt{trees} and \texttt{deer}, grounding the linkage to the act of {\it consuming} and reasoning about the effects of consumption on the populations of flora and fauna.

In order to evaluate the task of reasoning about the semantics of diagrams, we study semantic interpretation of diagrams in the context of diagram question answering. This is inspired by evaluation paradigms in school education systems and the recent progress in visual and textual question answering. Studying semantic interpretation of diagrams in the context of question answering also provides a well-defined problem definition, evaluation criteria, and metrics. 

\begin{figure}[t]
\centering
\includegraphics[width=27pc]{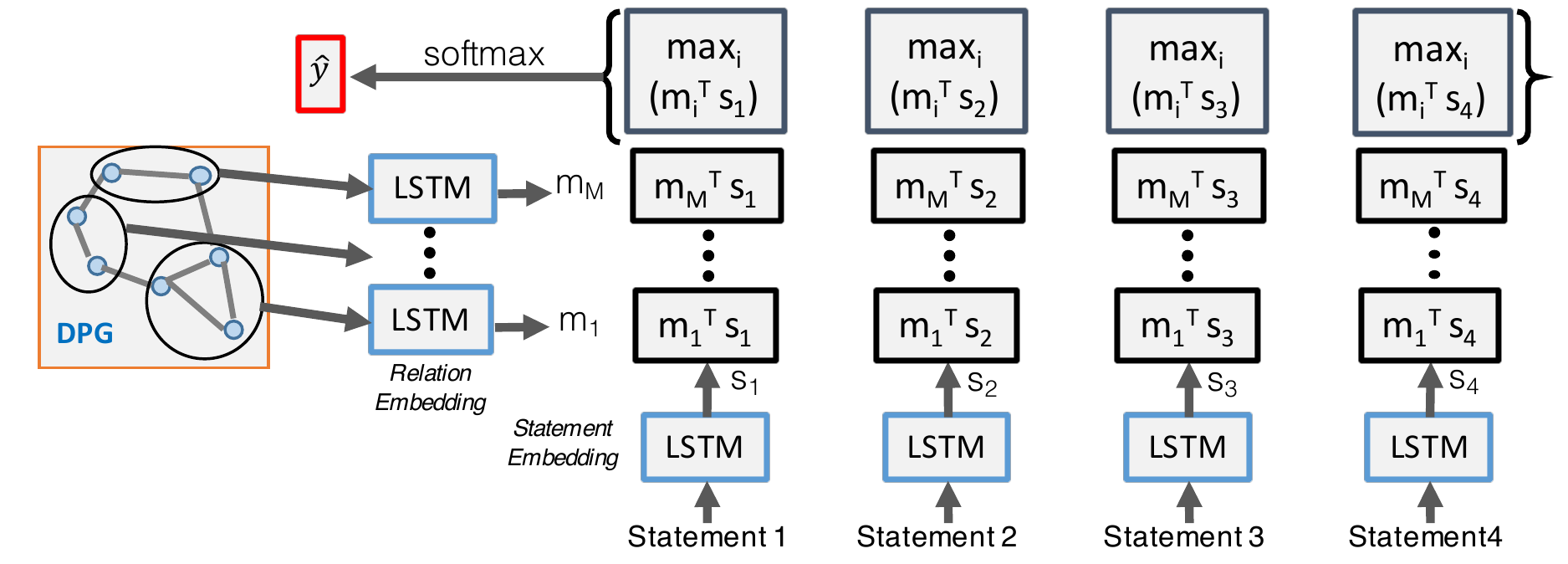}\vspace{-.2cm}
\caption{\small An overview of the \questionnet\ solution to diagram question answering. The network encodes the DPG into a set of facts, learns to attend on the most relevant fact, given a question and then answers the question.}
\label{fig:qaNetwork}
\end{figure}

\vspace{.1cm}
\noindent{\bf Diagram Question Answering.}
A diagram question consists of a diagram $d$ in raster graphics, a question sentence $q$, and multiple choices $\{c_1\ldots c_4\}$. The goal of question answering is to select a single correct choice $c_k$ given $d$ and $q$ (example questions in Figure~\ref{fig:qaqual}.) 

We design a neural network architecture (called \questionnet) to answer diagrammatic questions. The main intuition of the network is to encode the DPG into a set of facts and learn an attention model to find the closest fact to the question. For example, Figure~\ref{fig:qaqual} shows facts that \questionnet\ has selected to answer   questions. More formally, \questionnet\ consists of the following components: 

\noindent (a) a question embedding module that takes the question $q$ and a choice $c_k, k\in\{1\ldots4\}$ to build a statement $s_k$ and uses an LSTM to learn a $d$-dimensional embedding of the statement $s_k\in \mathbb{R}^d$; (b) a diagram embedding module that takes the DPG, extracts $M$ relations $m_i, i\in\{1\ldots M\}$ from DPG, and uses an LSTM to learn a $d$-dimensional embedding of diagram relations $m_i\in \mathbb{R}^d$; (c) an attention module that learns to attend to the relevant diagram relations by selecting the best statement choice that has a high similarity with  the relevant diagram relations. For every statement $s_k$, our model computes a probability distribution over statement choices by feeding the best similarity scores between statements and diagram relations through a softmax layer. 
\begin{equation*}
    \gamma_k=\max_i{s_k^T\cdot m_i}, \ \ \ \  \hat{y}=\text{softmax}_k(\gamma_k)=\frac{\exp(\gamma_k)}{\sum_{k'}\exp(\gamma_{k'})}
\end{equation*}
We use cross entropy loss to train our model: $  L(\theta)=H(y,\hat{y})=-\sum_ky_k\log \hat{y}_k$. More details about the parameters can be found in Section~\ref{sec:dqaexperiment}.

\section{Dataset} \label{Dataset}
We build a new dataset (named AI2 Diagrams (\datasetname)), to evaluate the task of diagram interpretation. \datasetname\ is comprised of more than 5,000 diagrams representing topics from grade school science, each annotated with constituent segmentations, their relationships to each other and their relationships to the diagram canvas. In total, \datasetname\ contains annotations for more than 118K constituents and 53K relationships. The dataset is also comprised of more than 15000 multiple choice questions associated to the diagrams. We divide the \datasetname\ dataset into a train set with 4000 images and a blind test set with 1000 images and report our numbers on this blind test set.

\begin{figure}[t]
\centering
\includegraphics[width=26pc]{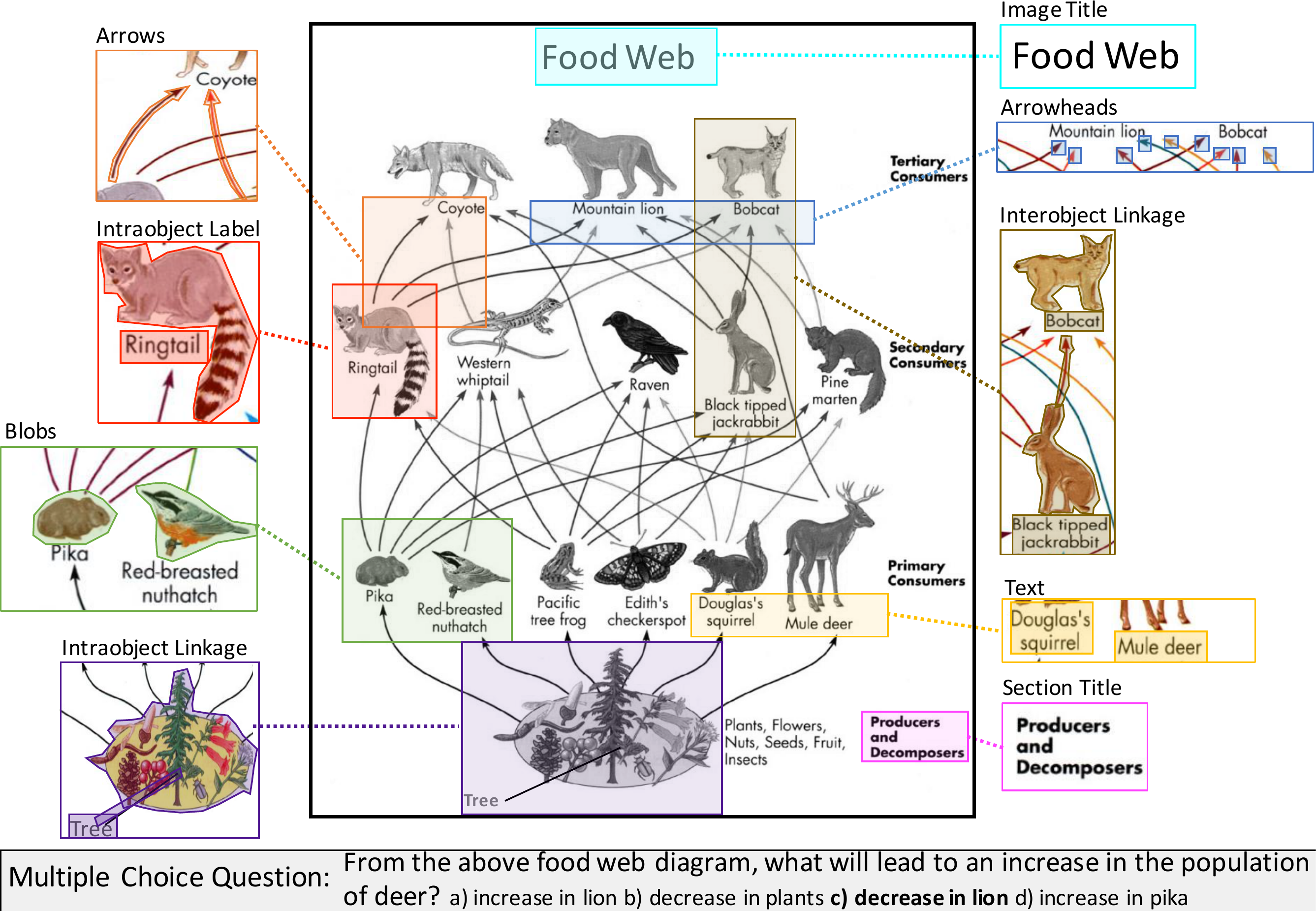}\vspace{-.2cm}
\caption{\small An image from the \datasetname\ dataset showing some of its rich annotations and a multiple choice question.}
\label{fig:datasetComplexity}
\end{figure}

The images are collected by scraping Google Image Search with seed terms derived from the chapter titles in Grade 1 - 6 science textbooks. Each image is annotated using Amazon Mechanical Turk (AMT). Annotating each image with rich annotations such as ours, is a rather complicated task and must be broken down 
into several phases to maximize the level of agreement obtained from turkers. Also, these phases need to be carried out sequentially to avoid conflicts in the annotations. The phases involve (1) annotating the four low-level constituents, (2) categorizing the text boxes into one of four categories: relationship with the canvas, relationship with a diagrammatic element, intra-object relationship and inter-object relationship, (3) categorizing the arrows into one of three categories: intra-object relationship, inter-object relationship or neither, (4) labelling intra-object linkage and inter-object linkage relationships. For this step, we display arrows to turkers and have them choose the origin and destination constituents in the diagram, (5) labelling intra-object label, intra-object region label and arrow descriptor relationships. For this purpose, we display text boxes to turkers and have them choose the constituents related to it, and finally (6) multiple choice questions with answers, representing grade school science questions are then obtained for each image using AMT. Figure~\ref{fig:datasetComplexity} shows some of the rich annotations obtained for an image in the dataset along with one of its associated multiple choice questions.

\section{Experiments} \label{sec:experiments}
We describe methods used to generate constituent and relationship proposals and show evaluations of our methods in generating proposals versus several baselines. We also evaluate our introduced model \lstmnetwork\  for syntactic parsing of diagrams that forms DPGs and compare it to several baseline approaches. Finally, we evaluate the proposed diagram question answering model \questionnet\ and compare with standard visual question answering approaches. In each section, we also describe the hyperparameters, features, and the baselines. 

\subsection{Generating Constituent Proposals}\label{sec:proposal1}

\noindent \textbf{Diagram Canvas:}
A diagram consists of multiple constituents overlaid onto a canvas, which may be uniformly colored, textured or have a blended image. We classify every pixel in the diagram into canvas vs constituent. We build non-parametric kernel density estimates (KDE) in RGB, texture and entropy spaces to generate features for a Random Forest (RF) classifier with 100 trees to obtain an Average Precision (AP) of 0.9142.

\begin{figure}[t]
\centering
\subfigure[Blobs]{
\includegraphics[width=0.22\textwidth]{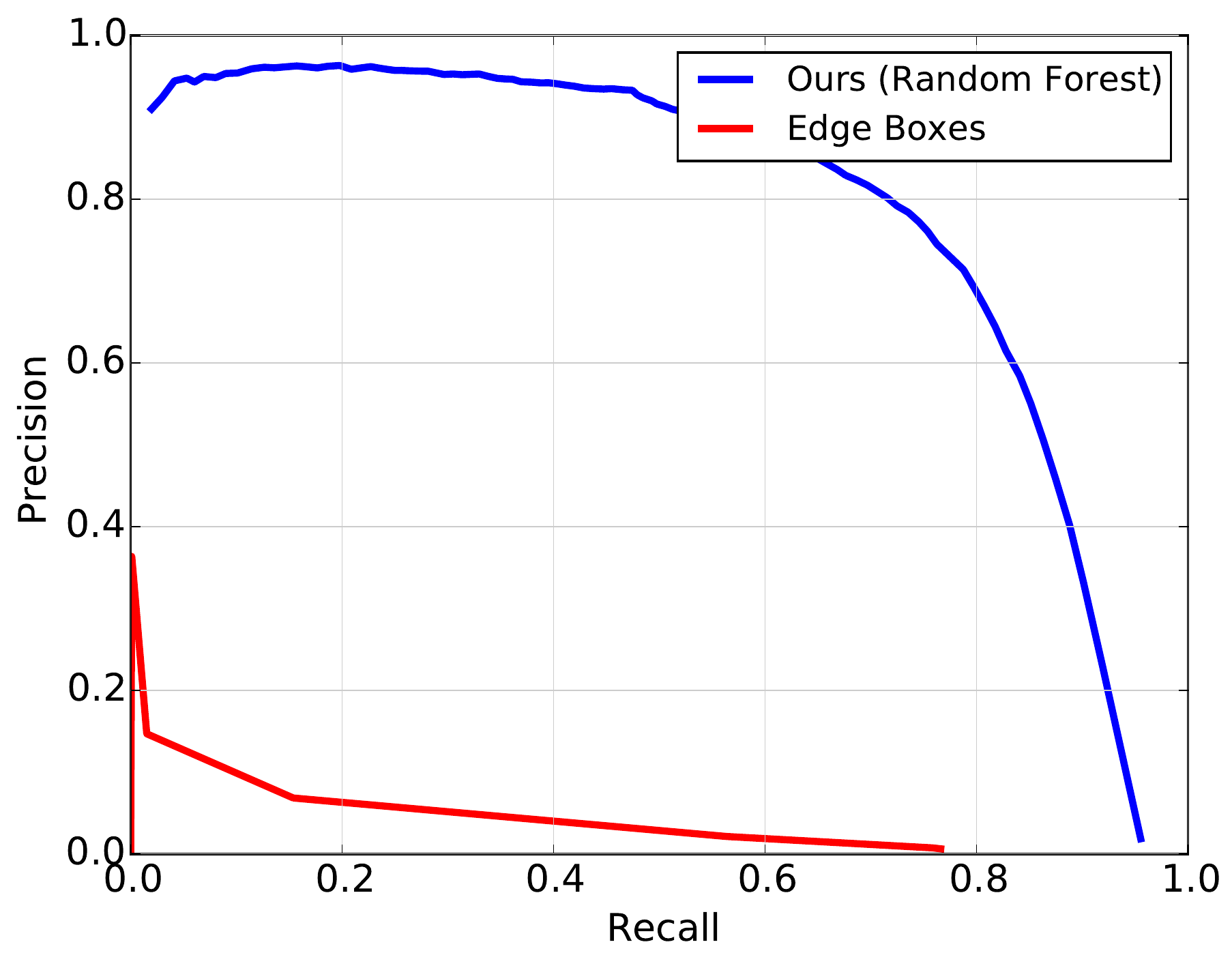}
\label{fig:blobsPR}}
\subfigure[Arrow Tails]{
\includegraphics[width=0.22\textwidth]{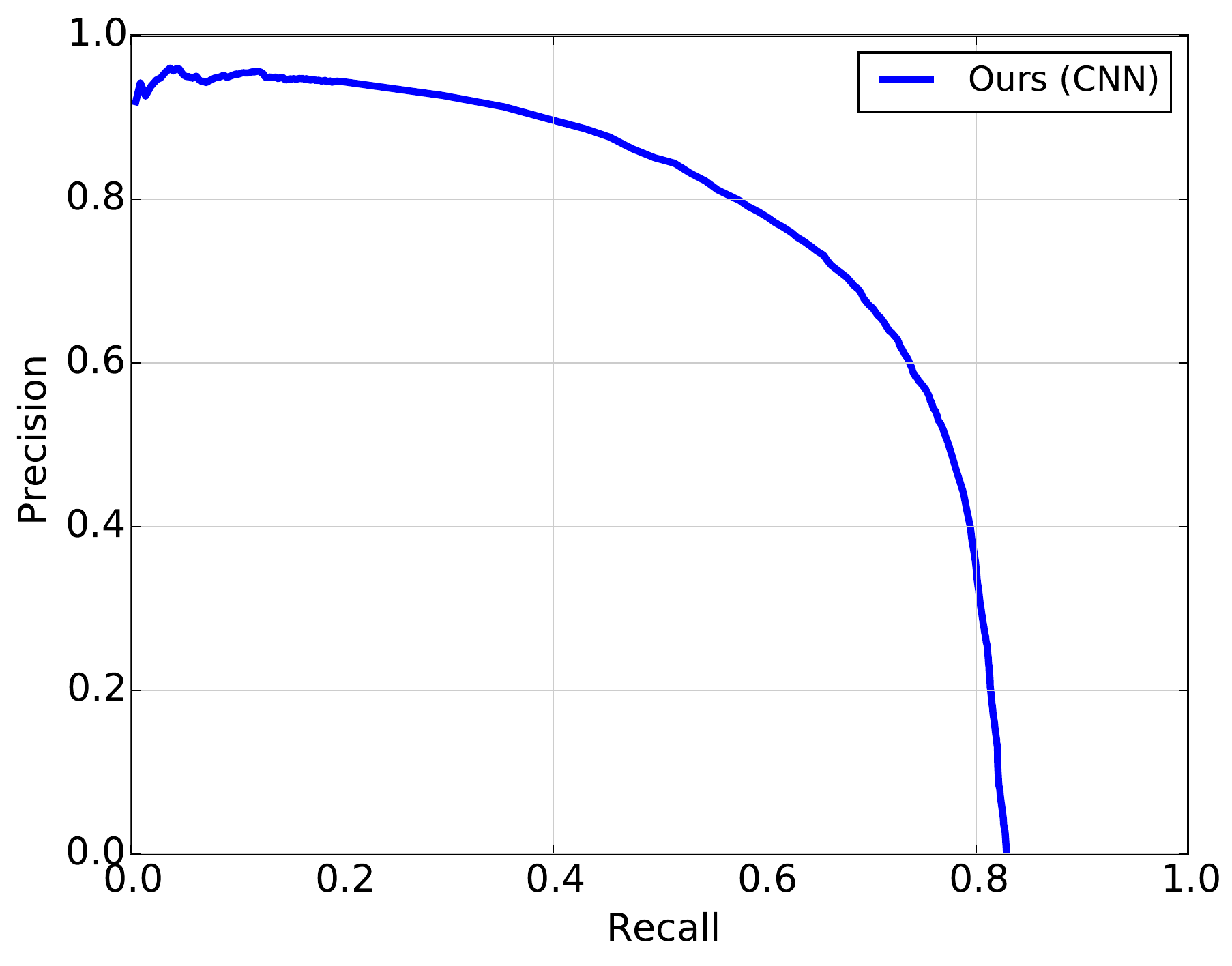}
\label{fig:arrowsPR}}
\subfigure[Arrow Heads]{
\includegraphics[width=0.22\textwidth]{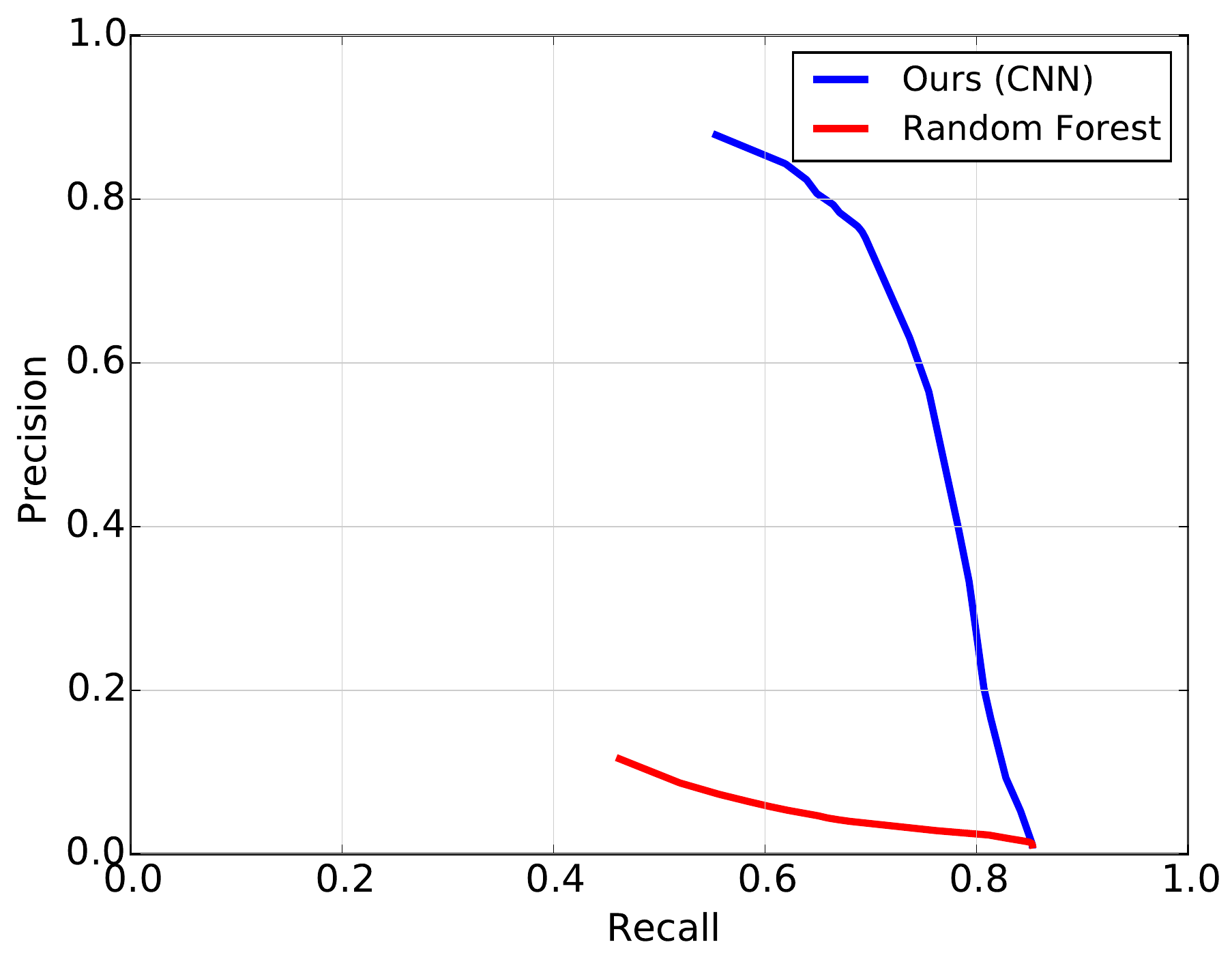}
\label{fig:arrowHeadsPR}}
\subfigure[Relationships]{
\includegraphics[width=0.22\textwidth]{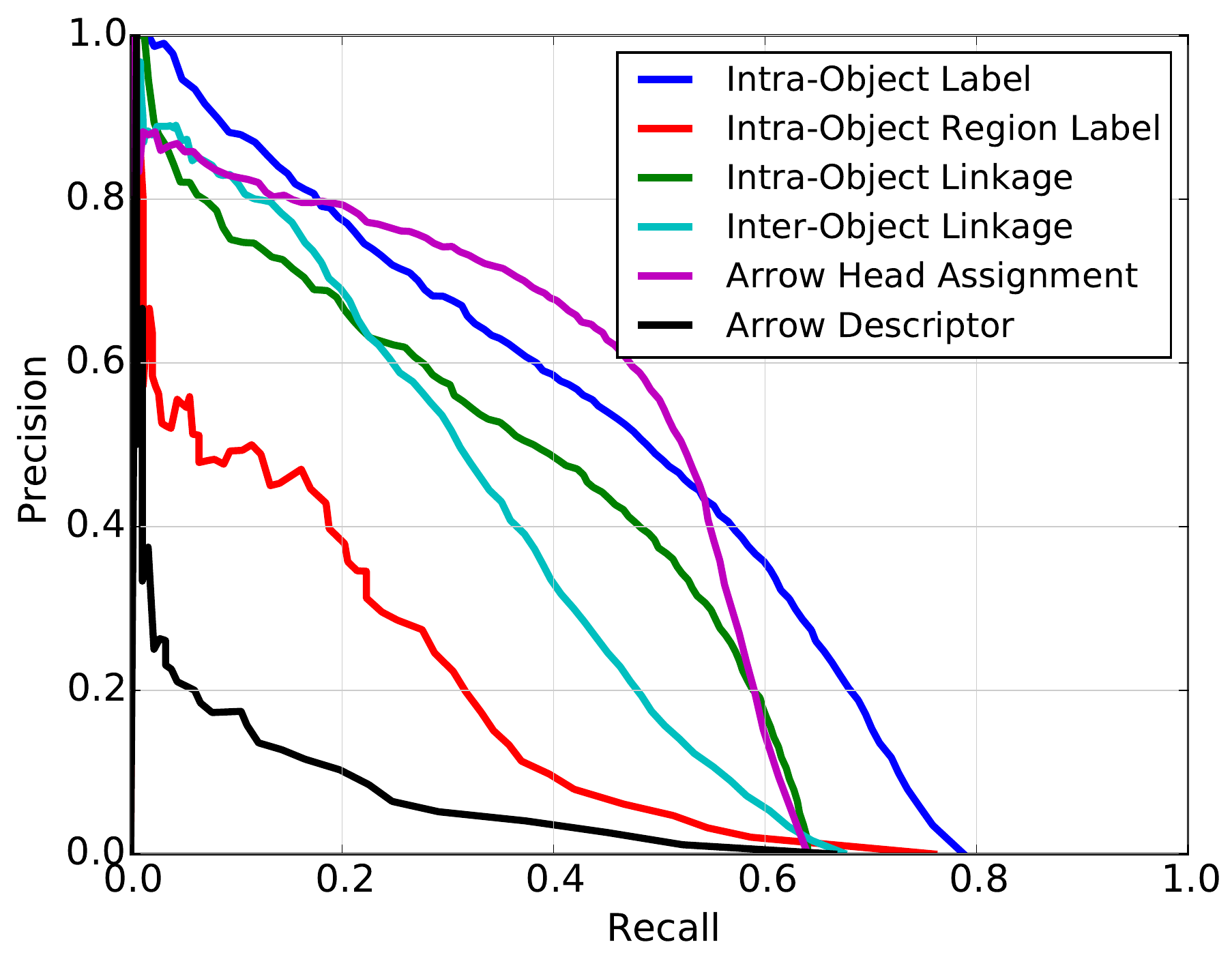}
\label{fig:relPR}} \vspace{-.2cm}
\caption{\small Precision Recall curves for constituent and relationship proposal generators.}
\label{fig:result}
\end{figure}

\vspace{.1cm}
\noindent \textbf{Detecting blobs:} Blobs exhibit a large degree of variability in their size, shape and appearance in diagrams, making them challenging to model. We combine segments at multiple levels of a segmentation hierarchy, obtained using Multiscale Combinatorial Grouping (MCG) \cite{Arbelez2014MultiscaleCG} 
with segments produced using the canvas probability map to produce a set of candidates. Features capturing the location, size, central and Hu moments, etc. are provided to an RF classifier with 100 trees. 
\noindent \textit{Baselines. }
We evaluated several object proposal approaches including Edge Boxes~\cite{Zitnick2014EdgeBL}, Objectness~\cite{Alexe2012MeasuringTO} and Selective Search~\cite{Uijlings2013SelectiveSF}. Since these are designed to work on natural images, they do not provide good results on diagrams. We compare the RF approach to Edge Boxes, the most suitable of these methods, since it uses edge maps to propose objects and relies less on colors and gradients observed in natural images. {\bf Results.} Our approach produces a significantly higher AP of 0.7829 compared to 0.02 (Figure~\ref{fig:blobsPR}).

\vspace{.1cm}
\noindent \textbf{Detecting arrow tails: } Arrow tails are challenging to model since they are easily confused with other line segments present in the diagram and do not always have a corresponding arrow head to provide context. We generate proposal segments using a three pronged approach. We obtain candidates using the boundary detection method in \cite{Kokkinos2010HighlyAB}, 
Hough transforms and by detecting parallel curved edge segments in a canny edge map; and recursively merge proximal segments that exhibit a low residual when fit to a 2\textsuperscript{nd} degree polynomial. We then train a binary class Convolutional Neural Network (CNN) resembling the architecture of the VGG-16 model by \cite{Simonyan2014VeryDC}, with a fourth channel appended to the standard three channel RGB input. This fourth channel specifies the location of the arrow tail candidate smoothed with a Gaussian kernel of width 5. All filters except the ones for the fourth input channel at layer 1 are initialized from a publicly available VGG-16 model. The remaining filters are initialized with random values drawn from a Gaussian distribution. We use a batch size of 32 and a starting learning rate (LR) of 0.001. {\bf Results.} Figure~\ref{fig:arrowsPR} shows the PR curve for our model with an AP of 0.6748. We tend to miss arrows that overlap significantly with more than three other arrows in the image as well as very thick arrows that are confused for blobs.

\vspace{.1cm}
\noindent \textbf{Detecting arrow heads: } Arrow head proposals are obtained by a scanning window approach over 6 scales and 16 orientations. RGB pixels in each window undergo PCA followed by a 250 tree RF classifier. We then train a binary class CNN resembling the standard architecture of AlexNet \cite{Krizhevsky2012ImageNetCW} and initialize using a publicly available model. We use a batch size of 128 and a starting LR of 0.001. {\bf Results.} Figure~\ref{fig:arrowHeadsPR} shows the PR curves for our CNN model as well as the first pass RF model. We miss arrow heads which are extremely small and some which are present in poor quality images.

\vspace{.1cm}
\noindent \textbf{Detecting text: } We use an Optical Character Recognition (OCR) service \cite{microsoft2015Oxford} provided by Microsoft's Project Oxford to localize and recognize text in our diagrams. To improve the performance on single characters, we train a single character localizer using a CNN having the same architecture as AlexNet \cite{Krizhevsky2012ImageNetCW}. We use three training resources: (1) Chars74K (a character recognition dataset for natural images \cite{deCampos09}), (2) a character dataset obtained from vector PDFs of scientific publications and (3) a set of synthetic renderings of single characters. The localized bounding boxes are then recognized using Tesseract \cite{tesseract}. {\bf Results.} Using Tesseract end-to-end provides poor text localization results for diagrams with a 0.2 precision and a 0.46 recall. Our method improves the precision to 0.89 and recall to 0.75. Our false negatives comprise of vertically oriented and curved text, cursive fonts and unmerged multi-line blocks.

\subsection{Generating Relationship Proposals}\label{sec:proposal2}
Relationship categories are presented in Table~\ref{tab:relations}. 
Categories $\mathbb{R}_1$ through $\mathbb{R}_6$ relate two or more constituents with one another. We compute features capturing the spatial layout of the constituents with respect to one another as well as the diagram space and combine them with detection probabilities provided by the low level constituent models. A 100 trees RF classifier is trained for each category. At test time, we generate proposal relationships from the large combinatorial set of candidate constituents using a proximity based pruning scheme, which get classified by the RF model. Categories $\mathbb{R}_7$ through $\mathbb{R}_{10}$ relate a single constituent with the entire image. We model each category using a non parametric Kernel Density Estimate (KDE) in X,Y space. At test time, every candidate text detection is passed through the KDE models to obtain a relationship probability. {\bf Results.} Figure~\ref{fig:relPR} shows the PR curves for the relationships built using the RF classifier. The AP for several of relationships is low, owing to the inherent ambiguity in classifying relationships using local spatial decisions. 

\begin{table}[h]
\centering 
\begin{tabular}{ccc}
    \centering
    \begin{tabular}{|c|c|}
    \hline
        Method & JIG Score  \\ \hline
        \textsc{Greedy Search}&  28.96  \\ \hline
        \textsc{A* Search} & 41.02  \\ \hline
        \lstmnetwork & \textbf{51.45}  \\
        \hline
    \end{tabular}
& \ \ \ \ \ \ \ \  &
    \begin{tabular}{|c|c|c|}
    \hline
         Method & Training set  & Accuracy \\ \hline
         VQA  & VQA & 29.06 \\ \hline
         VQA & \datasetname & 32.90 \\ \hline
         \questionnet  & \datasetname & \textbf{38.47}\\
         \hline
    \end{tabular}
\end{tabular}
    \caption{\small (left) Syntactic parsing results, (right) Question answering results}
    \label{tab:dpgEval}
\end{table}

\subsection{Syntactic Parsing: DPG Inference}\label{sec:lstmexperiment}

\noindent \textbf{Our model DSDP-Net: }
The introduced \lstmnetwork~(depicted in Figure~\ref{fig:lstmNetwork}) consists of a 2 layer stacked LSTM with each layer having a hidden state of dimensionality 512. The LSTM is preceded by two fully connected layers with an output dimensionality of 64 and a Rectified Linear Unit (ReLu)~\cite{Nair2010RectifiedLU} activation function each. The LSTM is proceeded by a fully connected layer with a softmax activation function. This network is trained using RMSProp~\cite{rmsPropHinton} to optimize the cross-entropy loss function. The initial learning rate is set to 0.0002. 

Each candidate relationship is represented as a 92 dimensional feature vector that includes features for each constituent in the relationship (normalized x,y coordinates, detection score, overlap ratio with higher scoring candidates and the presence of this constituent in relationships presented to the network at prior time-steps) and features describing the relationship itself (relationship score and the presence of tuples of candidates in relationships presented to the network at previous time steps). We sample 100 relationship sequences per training image to generate roughly 400000 training samples. At test time, relationships are presented to the network in sorted order, based on their detection scores.

\vspace{.1cm}
\noindent \textbf{Baselines: }
\textsc{Greedy Search: }
The first baseline is a greedy algorithm whereby nodes and edges are greedily added to the DPG using their proposal scores. It uses an {\it exit} model as a stopping condition. The exit model is trained to score DPGs based on their distance to the desired completed DPG. To train the exit model, we use features capturing the quality, coverage, redundancy and structure of the nodes and edges in the DPGs and use 100 tree RF models. 

\textsc{A* Search}:
The second baseline is an A* search, which starts from an empty DPG and sequentially adds nodes and edges according to a cost. We improve upon the greedy algorithm by training a RF model that utilizes local and contextual cues to rank available constituents and relationships. The cost function for each potential step is a linear combination of this RF model's score and the distance of the resultant DPG to the desired complete DPG. In order to model the distance function, we use the same exit model as before to approximate the distance from the goal. 

\begin{figure}[t]
    \centering
    \includegraphics[scale=0.6]{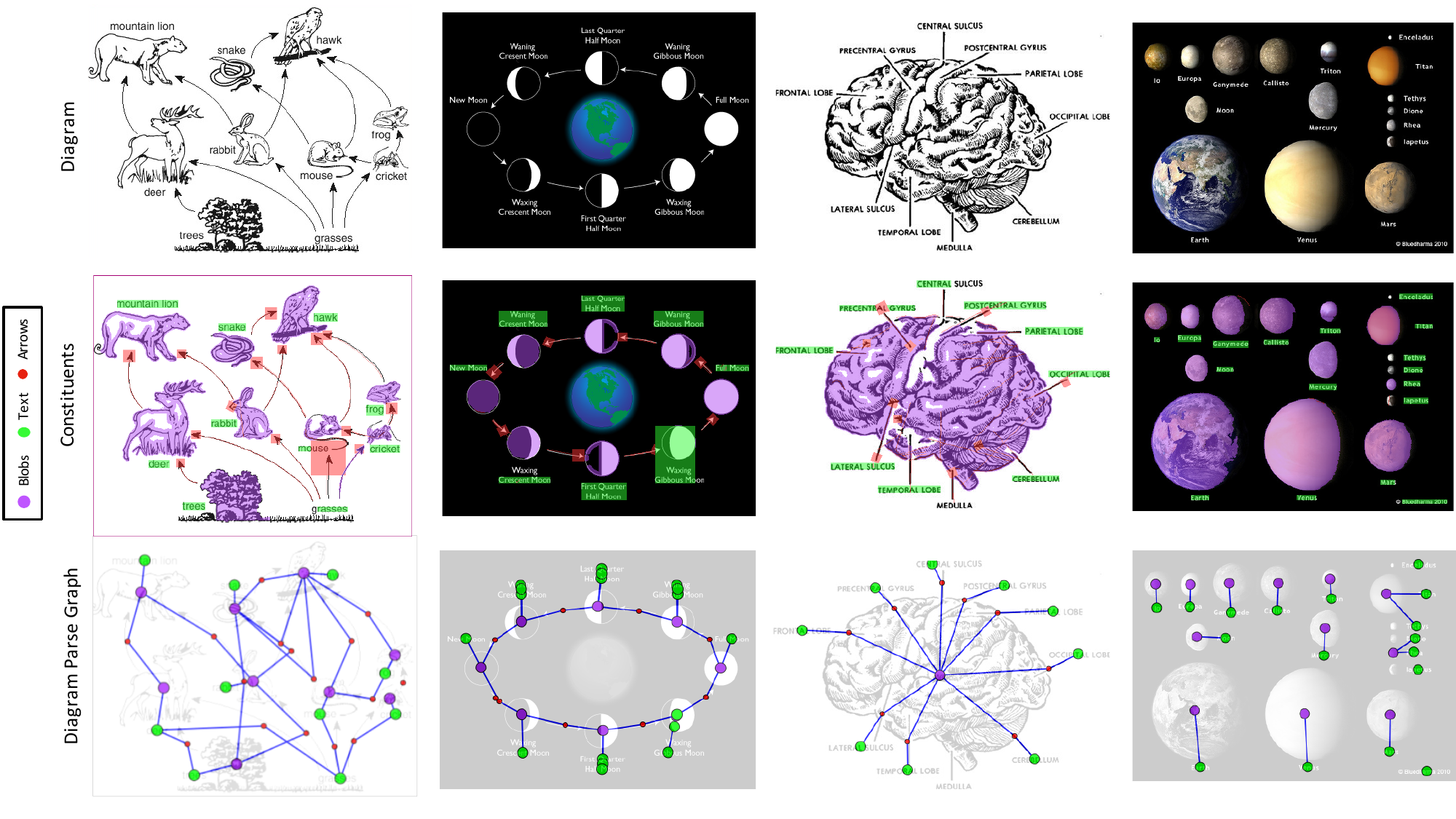}\vspace{-.2cm}
    \caption{\small Inferred DPGs using \lstmnetwork. The first row shows the diagram, the second row shows the constituent segmentations and the third row shows the inferred DPGs.}
    \label{fig:dpgs}
\end{figure}

\textsc{Direct Regression: }
We also trained a CNN to directly regress the DPG, akin to YOLO~\cite{Redmon2015YouOL}. This generated no meaningful results on our dataset.

\noindent {\bf Evaluation.} To evaluate these methods, we compute the Jaccard Index between the sets of nodes and edges in our proposed DPG and and the ground truth DPG. We refer to this metric by the Jaccard Index for Graphs (JIG) score. The Jaccard Index, which measures similarity between finite sample sets, is defined as the size of the intersection divided by the size of the union of the sample sets. 

\noindent {\bf Results.} Table~\ref{tab:dpgEval}({\it left}) shows the mean JIG scores, computed over the test set for each method. The \lstmnetwork\ method outperforms both the \textsc{Greedy Search} and \textsc{A* search} by a considerable margin. This shows the importance of our sequential formulation to use LSTMs for adding relationships to form DPGs. Figure~\ref{fig:dpgs} shows qualitative examples of inferred DPGs using \lstmnetwork.

\subsection{Diagram Question Answering}\label{sec:dqaexperiment}
\noindent{\bf Our model DQA-Net:} \questionnet\ uses GloVe~\cite{pennington2014glove} model pre-trained on 6B tokens (Wikipedia 2014) to map each word to a 300D vector.
The LSTM units have a single layer, $50$ hidden units, and forget bias of $2.5$.
We place a single 50-by-300 FC layer between the word vectors and the LSTM units.
The LSTM variables in all sentence embeddings (relation and statement) are shared.
The loss function is optimized with stochastic gradient descent with the batch size of $100$.
Learning rate starts at $0.01$ and decays by $0.5$ in every $25$ epochs, for $100$ epochs in total.

\noindent{\bf Baselines.} We use the best model (LSTM Q+I) from~\cite{antol2015vqa} as the baseline, which consists of an LSTM module for statement embedding and a CNN module for diagram image embedding.
In the LSTM module, we use the same setup as \questionnet, translating question-answer pairs to statements and obtaining 50D vector for each statement. 
In the CNN module, we obtain 4096D vector from the last layer of pre-trained VGG-19 model~\cite{Simonyan14c} for each diagram image. Each image vector is transformed to a 50D vector by a single 50-by-4096 FC layer. 
We then compute the dot product between each statement embedding and the transformed 50D image vector, 
followed by a softmax layer. We use cross entropy loss and the same optimization techniques as in \questionnet.

\begin{figure}[t]
    \centering
    \includegraphics[scale=0.49]{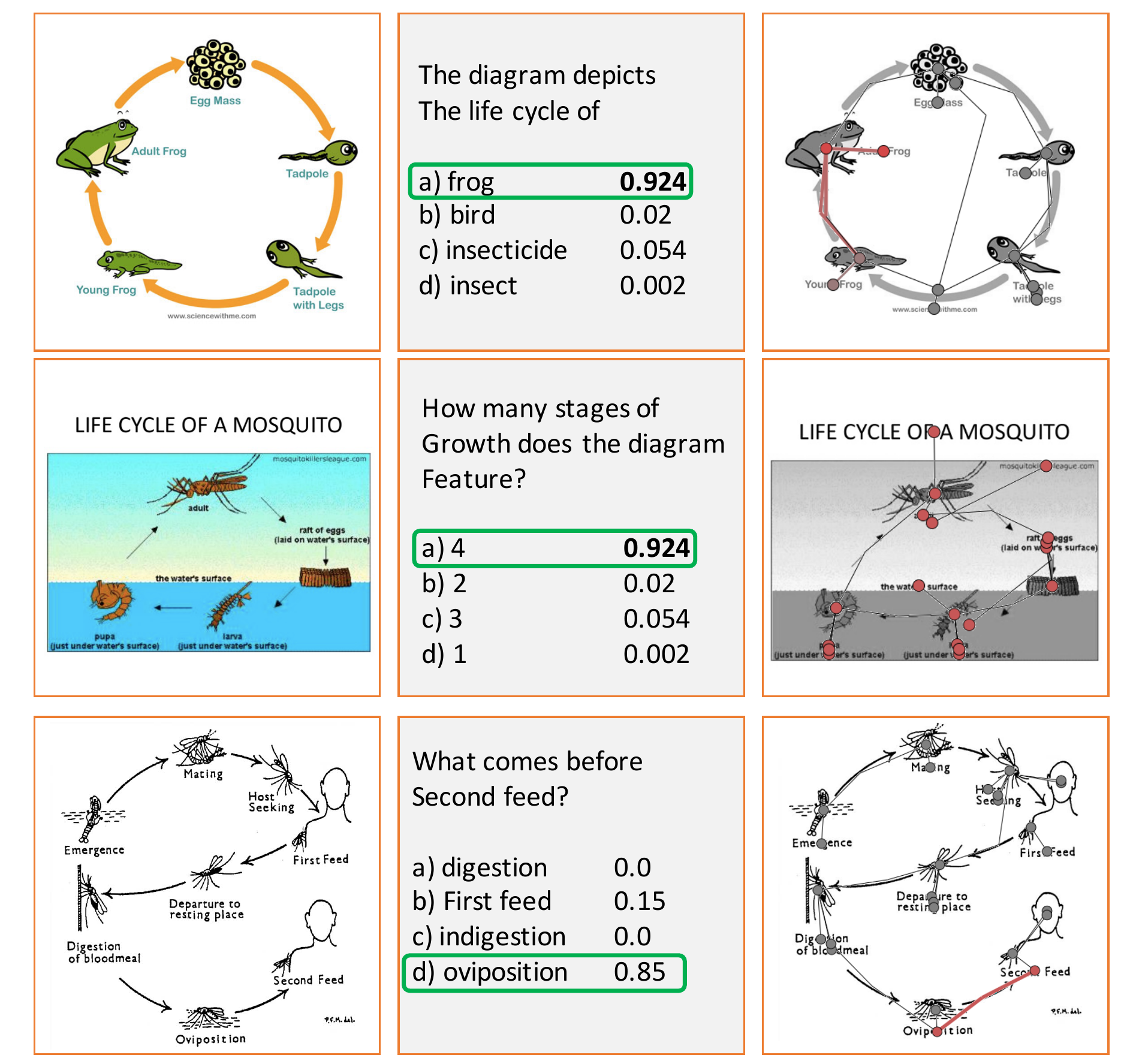}\vspace{-.3cm}
    \caption{\small Sample question answering results using \questionnet. The second column shows the answer chosen and the third column shows the nodes and edges in the DPG that \questionnet\ decided to attend to (indicated by red highlights).}
    \label{fig:qaqual}
\end{figure}

\noindent {\bf Results.} Table~\ref{tab:dpgEval} ({\it right}) reports the accuracy of different methods for question answering on the test set. \questionnet\ method outperforms the baseline both when it is trained on the VQA dataset as well as the on \datasetname. This shows that the DPG effectively encodes high-level semantics of the diagrams, which are required to answer \datasetname\  questions. Figure~\ref{fig:qaqual} shows examples of correctly answered questions by \questionnet.

\subsection {Libraries}
We use Keras~\cite{chollet2015keras} to build our constituent CNN models and the \lstmnetwork\ network, TensorFlow~\cite{tensorflow2015-whitepaper} to build our \questionnet\ network and Scikit-learn~\cite{scikit-learn} to build our Random Forest models.
\vspace{-.3cm}
\section{Conclusion} \vspace{-.1cm}
We introduced the task of diagram interpretation and reasoning.  We proposed \lstmnetwork\ to parse diagrams and reason about the global context of a diagram using our proposed representation called DPG. DPGs encode most useful syntactic  information depicted in the diagram. Moreover, we introduced \questionnet\ that learns to answer diagram questions by attending to diagram relations encoded with DPGs. The task of diagram question answering is a well-defined task to evaluate capabilities of different systems in semantic interpretation of diagrams and reasoning. Our experimental results show improvements of \lstmnetwork\ in parsing diagrams compared to strong baselines. Moreover, we show that \questionnet\ outperforms standard VQA techniques in diagram question answering.

Diagram interpretation and reasoning raises new research questions that goes beyond natural image understanding. We release \datasetname\ and our baselines to facilitate further research in diagram understanding and reasoning. Future work involves incorporating diagrammatic and commonsense knowledge in DQA. 

\bibliographystyle{splncs}
\bibliography{diagramUnderstanding2016Refs}

\begin{thebibliography}{10}

\bibitem{Srihari1994ComputationalMF}
Srihari, R.K.:
\newblock Computational models for integrating linguistic and visual
  information: A survey.
\newblock Artif. Intell. Rev. \textbf{8} (1994)  349--369

\bibitem{Ferguson98tellingjuxtapositions}
Ferguson, R.W., Forbus, K.D.:
\newblock Telling juxtapositions: Using repetition and alignable difference in
  diagram understanding.
\newblock Advances in Analogy Research (1998)

\bibitem{Watanabe1998DiagramUU}
Watanabe, Y., Nagao, M.:
\newblock Diagram understanding using integration of layout information and
  textual information.
\newblock In: ACL. (1998)

\bibitem{ogorman1997dia}
O'Gorman, L., Kasturi, R.:
\newblock Document Image Analysis.
\newblock IEEE Computer Society Executive Briefings (1997)

\bibitem{Futrelle2003ExtractionLA}
Futrelle, R.P., Shao, M., Cieslik, C., Grimes, A.E.:
\newblock Extraction, layout analysis and classification of diagrams in pdf
  documents.
\newblock In: ICDAR. (2003)

\bibitem{Seo2014DiagramUI}
Seo, M.J., Hajishirzi, H., Farhadi, A., Etzioni, O.:
\newblock Diagram understanding in geometry questions.
\newblock In: AAAI. (2014)

\bibitem{von2002language}
von Engelhardt, J.:
\newblock The language of graphics: A framework for the analysis of syntax and
  meaning in maps, charts and diagrams.
\newblock Yuri Engelhardt (2002)

\bibitem{Zitnick2013BringingSI}
Zitnick, C.L., Parikh, D.:
\newblock Bringing semantics into focus using visual abstraction.
\newblock In: CVPR. (2013)

\bibitem{Zhang2015YinAY}
Zhang, P., Goyal, Y., Summers-Stay, D., Batra, D., Parikh, D.:
\newblock Yin and yang: Balancing and answering binary visual questions.
\newblock CoRR \textbf{abs/1511.05099} (2015)

\bibitem{Vedantam2015LearningCS}
Vedantam, R., Lin, X., Batra, T., Zitnick, C.L., Parikh, D.:
\newblock Learning common sense through visual abstraction.
\newblock In: ICCV. (2015)

\bibitem{Antol2014ZeroShotLV}
Antol, S., Zitnick, C.L., Parikh, D.:
\newblock Zero-shot learning via visual abstraction.
\newblock In: ECCV. (2014)

\bibitem{Tu2003ImagePU}
Tu, Z., Chen, X., Yuille, A.L., Zhu, S.C.:
\newblock Image parsing: Unifying segmentation, detection, and recognition.
\newblock In: CLOR. (2003)

\bibitem{Zhu2006ASG}
Zhu, S.C., Mumford, D.:
\newblock A stochastic grammar of images.
\newblock Foundations and Trends in Computer Graphics and Vision \textbf{2}
  (2006)  259--362

\bibitem{Martinovic2013BayesianGL}
Martinovic, A., Gool, L.J.V.:
\newblock Bayesian grammar learning for inverse procedural modeling.
\newblock In: CVPR. (2013)

\bibitem{Pirsiavash2014ParsingVO}
Pirsiavash, H., Ramanan, D.:
\newblock Parsing videos of actions with segmental grammars.
\newblock In: CVPR. (2014)

\bibitem{Choi2013UnderstandingIS}
Choi, W., Chao, Y.W., Pantofaru, C., Savarese, S.:
\newblock Understanding indoor scenes using 3d geometric phrases.
\newblock In: CVPR. (2013)

\bibitem{Socher2011ParsingNS}
Socher, R., Lin, C.C.Y., Ng, A.Y., Manning, C.D.:
\newblock Parsing natural scenes and natural language with recursive neural
  networks.
\newblock In: ICML. (2011)

\bibitem{WestonBCM15}
Weston, J., Bordes, A., Chopra, S., Mikolov, T.:
\newblock Towards ai-complete question answering: {A} set of prerequisite toy
  tasks.
\newblock CoRR \textbf{abs/1502.05698} (2015)

\bibitem{RichardsonBR13}
Richardson, M., Burges, C.J.C., Renshaw, E.:
\newblock Mctest: {A} challenge dataset for the open-domain machine
  comprehension of text.
\newblock In: Proceedings of the 2013 Conference on Empirical Methods in
  Natural Language Processing, {EMNLP} 2013, 18-21 October 2013, Grand Hyatt
  Seattle, Seattle, Washington, USA, {A} meeting of SIGDAT, a Special Interest
  Group of the {ACL}. (2013)  193--203

\bibitem{hermann2015teaching}
Hermann, K.M., Kocisky, T., Grefenstette, E., Espeholt, L., Kay, W., Suleyman,
  M., Blunsom, P.:
\newblock Teaching machines to read and comprehend.
\newblock In: Advances in Neural Information Processing Systems. (2015)
  1684--1692

\bibitem{sukhbaatar2015end}
Sukhbaatar, S., Weston, J., Fergus, R.,  et~al.:
\newblock End-to-end memory networks.
\newblock In: Advances in Neural Information Processing Systems. (2015)
  2431--2439

\bibitem{antol2015vqa}
Antol, S., Agrawal, A., Lu, J., Mitchell, M., Batra, D., Lawrence~Zitnick, C.,
  Parikh, D.:
\newblock Vqa: Visual question answering.
\newblock In: Proceedings of the IEEE International Conference on Computer
  Vision. (2015)  2425--2433

\bibitem{ren2015exploring}
Ren, M., Kiros, R., Zemel, R.:
\newblock Exploring models and data for image question answering.
\newblock In: Advances in Neural Information Processing Systems. (2015)
  2935--2943

\bibitem{ZhuGBF15}
Zhu, Y., Groth, O., Bernstein, M.S., Fei{-}Fei, L.:
\newblock Visual7w: Grounded question answering in images.
\newblock CoRR \textbf{abs/1511.03416} (2015)

\bibitem{andreas2016learning}
Andreas, J., Rohrbach, M., Darrell, T., Klein, D.:
\newblock Learning to compose neural networks for question answering.
\newblock arXiv preprint arXiv:1601.01705 (2016)

\bibitem{noh2015image}
Noh, H., Seo, P.H., Han, B.:
\newblock Image question answering using convolutional neural network with
  dynamic parameter prediction.
\newblock arXiv preprint arXiv:1511.05756 (2015)

\bibitem{horn1998visual}
Horn, R.:
\newblock Visual language: Global communication for the 21st century.
\newblock Century. Brainbridge Island, WA, MacroVU (1998)

\bibitem{card1999readings}
Card, S.K., Mackinlay, J.D., Shneiderman, B.:
\newblock Readings in information visualization: using vision to think.
\newblock Morgan Kaufmann (1999)

\bibitem{twyman1979schema}
Twyman, M.:
\newblock A schema for the study of graphic language (tutorial paper).
\newblock In: Processing of visible language.
\newblock Springer (1979)  117--150

\bibitem{Hochreiter1997LongSM}
Hochreiter, S., Schmidhuber, J.:
\newblock Long short-term memory.
\newblock Neural Computation \textbf{9} (1997)  1735--1780

\bibitem{Arbelez2014MultiscaleCG}
Arbelaez, P.A., Pont-Tuset, J., Barron, J.T., Marques, F., Malik, J.:
\newblock Multiscale combinatorial grouping.
\newblock In: CVPR. (2014)

\bibitem{Zitnick2014EdgeBL}
Zitnick, C.L., r, P.D.:
\newblock Edge boxes: Locating object proposals from edges.
\newblock In: ECCV. (2014)

\bibitem{Alexe2012MeasuringTO}
Alexe, B., Deselaers, T., Ferrari, V.:
\newblock Measuring the objectness of image windows.
\newblock IEEE Trans. Pattern Anal. Mach. Intell. \textbf{34} (2012)
  2189--2202

\bibitem{Uijlings2013SelectiveSF}
Uijlings, J.R.R., van~de Sande, K.E.A., Gevers, T., Smeulders, A.W.M.:
\newblock Selective search for object recognition.
\newblock International Journal of Computer Vision \textbf{104} (2013)
  154--171

\bibitem{Kokkinos2010HighlyAB}
Kokkinos, I.:
\newblock Highly accurate boundary detection and grouping.
\newblock In: CVPR. (2010)

\bibitem{Simonyan2014VeryDC}
Simonyan, K., Zisserman, A.:
\newblock Very deep convolutional networks for large-scale image recognition.
\newblock CoRR \textbf{abs/1409.1556} (2014)

\bibitem{Krizhevsky2012ImageNetCW}
Krizhevsky, A., Sutskever, I., Hinton, G.E.:
\newblock Imagenet classification with deep convolutional neural networks.
\newblock In: NIPS. (2012)

\bibitem{microsoft2015Oxford}
Microsoft:
\newblock Project oxford.
\newblock \url{https://www.projectoxford.ai/}

\bibitem{deCampos09}
de~Campos, T.E., Babu, B.R., Varma, M.:
\newblock Character recognition in natural images.
\newblock In: Proceedings of the International Conference on Computer Vision
  Theory and Applications, Lisbon, Portugal. (February 2009)

\bibitem{tesseract}
Tesseract:
\newblock Open source ocr engine.
\newblock \url{https://github.com/tesseract-ocr/tesseract}

\bibitem{Nair2010RectifiedLU}
Nair, V., Hinton, G.E.:
\newblock Rectified linear units improve restricted boltzmann machines.
\newblock In: ICML. (2010)

\bibitem{rmsPropHinton}
Tieleman, T., Hinton, G.E.:
\newblock Lecture 6.5-rmsprop: Divide the gradient by a running average of its
  recent magnitude.
\newblock COURSERA: Neural Networks for Machine Learning (2012)

\bibitem{Redmon2015YouOL}
Redmon, J., Divvala, S.K., Girshick, R.B., Farhadi, A.:
\newblock You only look once: Unified, real-time object detection.
\newblock CoRR \textbf{abs/1506.02640} (2015)

\bibitem{pennington2014glove}
Pennington, J., Socher, R., Manning, C.D.:
\newblock Glove: Global vectors for word representation.
\newblock In: Empirical Methods in Natural Language Processing (EMNLP). (2014)
  1532--1543

\bibitem{Simonyan14c}
Simonyan, K., Zisserman, A.:
\newblock Very deep convolutional networks for large-scale image recognition.
\newblock CoRR \textbf{abs/1409.1556} (2014)

\bibitem{chollet2015keras}
Chollet, F.:
\newblock Keras.
\newblock \url{https://github.com/fchollet/keras} (2015)

\bibitem{tensorflow2015-whitepaper}
Abadi, M., Agarwal, A., Barham, P., Brevdo, E., Chen, Z., Citro, C., Corrado,
  G.S., Davis, A., Dean, J., Devin, M., Ghemawat, S., Goodfellow, I., Harp, A.,
  Irving, G., Isard, M., Jia, Y., Jozefowicz, R., Kaiser, L., Kudlur, M.,
  Levenberg, J., Man\'{e}, D., Monga, R., Moore, S., Murray, D., Olah, C.,
  Schuster, M., Shlens, J., Steiner, B., Sutskever, I., Talwar, K., Tucker, P.,
  Vanhoucke, V., Vasudevan, V., Vi\'{e}gas, F., Vinyals, O., Warden, P.,
  Wattenberg, M., Wicke, M., Yu, Y., Zheng, X.:
\newblock {TensorFlow}: Large-scale machine learning on heterogeneous systems
  (2015) Software available from tensorflow.org.

\bibitem{scikit-learn}
Pedregosa, F., Varoquaux, G., Gramfort, A., Michel, V., Thirion, B., Grisel,
  O., Blondel, M., Prettenhofer, P., Weiss, R., Dubourg, V., Vanderplas, J.,
  Passos, A., Cournapeau, D., Brucher, M., Perrot, M., Duchesnay, E.:
\newblock Scikit-learn: Machine learning in {P}ython.
\newblock Journal of Machine Learning Research \textbf{12} (2011)  2825--2830

\end{thebibliography}
\end{document}